\documentclass[letterpaper, 10 pt, journal, twoside]{IEEEtran}
\usepackage{amsmath, amssymb}
\usepackage{graphicx}
\usepackage{caption}
\usepackage{xcolor}
\usepackage[normalem]{ulem}
\usepackage{siunitx}
\usepackage[caption=false,font=footnotesize]{subfig}
\usepackage{titlesec}
\usepackage{multirow}
\usepackage{url}
\usepackage{hyperref}
\ifCLASSINFOpdf
\else
\fi

\begin{document}
\title{Dynamically-Consistent Trajectory Optimization for Legged Robots via Contact Point Decomposition}

% \author{Michael~Shell,~\IEEEmembership{Member,~IEEE,}
%         John~Doe,~\IEEEmembership{Fellow,~OSA,}
%         and~Jane~Doe,~\IEEEmembership{Life~Fellow,~IEEE}% <-this % stops a space
% \thanks{M. Shell was with the Department
% of Electrical and Computer Engineering, Georgia Institute of Technology, Atlanta,
% GA, 30332 USA e-mail: (see http://www.michaelshell.org/contact.html).}% <-this % stops a space
% \thanks{J. Doe and J. Doe are with Anonymous University.}% <-this % stops a space
% \thanks{Manuscript received April 19, 2005; revised August 26, 2015.}}
\author{Sangmin Kim$^{1}$, Hajun Kim$^{1}$, Gijeong Kim$^{1}$, Min-Gyu Kim$^{1}$ and Hae-Won Park$^{1}$, \textit{Member, IEEE}%
\thanks{Manuscript received: March, 31, 2025; Revised July, 14, 2025; Accepted October, 27, 2025.}%Use only for final RAL version
\thanks{This paper was recommended for publication by Editor Olivier Stasse upon evaluation of the Associate Editor and Reviewers' comments.
This work was supported by the Technology Innovation Program(or Industrial Strategic Technology Development Program-Robot Industry Technology Development)(RS-2024-00427719, Dexterous and Agile Humanoid Robots for Industrial Applications) funded By the Ministry of  Trade Industry \& Energy(MOTIE, Korea).
}% <-this % stops a space
\thanks{$^{1}$All authors are with the Humanoid Robot Research Center, Korea
Advanced Institute of Science and Technology, Daejeon 34141, Korea.
        {\tt\small haewonpark@kaist.ac.kr}}%
\thanks{Digital Object Identifier (DOI): see top of this page.}
}

\markboth{IEEE Robotics and Automation Letters. Preprint Version. Accepted October, 2025}
{Kim \MakeLowercase{\textit{et al.}}: Dynamically-Consistent Trajectory Optimization for Legged Robots via Contact Point Decomposition} 




\maketitle

\begin{abstract}
To generate reliable motion for legged robots through trajectory optimization, it is crucial to simultaneously compute the robot's path and contact sequence, as well as accurately consider the dynamics in the problem formulation. 
In this paper, we present a phase-based trajectory optimization that ensures the feasibility of translational dynamics and friction cone constraints throughout the entire trajectory. Specifically, our approach leverages the superposition properties of linear differential equations to decouple the translational dynamics for each contact point, which operates under different phase sequences. Furthermore, we utilize the differentiation matrix of B{\'e}zier polynomials to derive an analytical relationship between the robot's position and force, thereby ensuring the consistent satisfaction of translational dynamics. Additionally, by exploiting the convex closure property of B{\'e}zier polynomials, our method ensures compliance with friction cone constraints. Using the aforementioned approach, the proposed trajectory optimization framework can generate dynamically reliable motions with various gait sequences for legged robots. We validate our framework using a quadruped robot model, focusing on the feasibility of dynamics and motion generation. The source code and additional materials are publicly available at our project page:
\url{https://sangmin11.github.io/Analytical_Dynamics_TO_Project/}.
\end{abstract}

% \begin{IEEEkeywords}
% IEEE, IEEEtran, journal, \LaTeX, paper, template.
% \end{IEEEkeywords}
\begin{IEEEkeywords}
Legged Robots, Optimization and Optimal Control, Multi-Contact Whole-Body Motion Planning and Control
\end{IEEEkeywords}

% \fi
\IEEEpeerreviewmaketitle

\section{Introduction}
% for IEEE journal papers produced under \LaTeX\ using
% IEEEtran.cls version 1.8b and later.
\IEEEPARstart{T}{o} generate versatile motions for legged robots, it is crucial to consider not only body movements but also contact timing and gait sequences \cite{mordatch2012discovery, posa2014direct, dai2014whole}. Trajectory Optimization (TO) with contact scheduling typically models contact modes at each support point and timestep as binary variables, reflecting the discrete nature of contact decisions. Some studies formulate this problem as a Mixed Integer Programming (MIP) problem \cite{ibanez2014emergence}, \cite{aceituno2017simultaneous}. However, MIP is NP-hard, and the solve time increases exponentially with the number of discrete transitions.

On the other hand, Contact Implicit Trajectory Optimization (CITO) approach implicitly incorporates hybrid contact modes, rather than using binary variables, into a nonlinear programming (NLP) formulation~\cite{wensing2023optimization}. Specifically, CITO can be broadly categorized into two approaches. One is to incorporate complementarity constraints derived from the Signorini condition, and the other is the Phase-Based Trajectory Optimization (PBTO) approach, which explicitly treats the swing and stance phase durations as optimization variables for each support point, rather than considering complementarity constraints.

CITO approaches based on complementarity constraints have been shown to generate a variety of contact-feasible motions~\cite{posa2014direct, yunt2006trajectory}.
In particular, recent works have achieved fast update rates suitable for Model Predictive Control (MPC) by leveraging the structural properties of trajectory optimization, while demonstrating successful hardware implementation \cite{aydinoglu2022real, kim2023contact, le2024fast}. However, due to the exponential number of possible mode combinations with respect to the number of support points at each timestep, the optimization landscape contains many poor local minima \cite{posa2014direct}. To guide the solver toward a desirable solution, relaxation is typically used in conjunction with appropriate reference trajectories \cite{le2024fast} or well-designed cost terms \cite{kim2023contact}. Nevertheless, these methods often focus on horizons in the range of a few tenths of a second.

PBTO \cite{winkler2018gait, ahn2021versatile} assumes that the contact sequence of each leg follows a sequential repetition of stance and swing phases, and contact patterns by adjusting the durations of these repeated phases. In particular, PBTO can handle horizons on the order of a few seconds and serve as reference trajectories for model-based controllers \cite{ahn2021versatile} or target motions for learning-based controllers~\cite{brakel2022learning, wu2023learning}.

However, although PBTO can generate motions with various gait sequences by adjusting phase durations, the framework lacks confirmation on whether the dynamics feasibility is ensured~\cite{winkler2018gait, ahn2021versatile}. For example, to ensure the dynamics feasibility, previous work \cite{winkler2018gait} applies dynamics constraints at specific nodes spaced at equal time intervals. In such cases, dynamic feasibility may not be guaranteed between the constrained nodes. This issue becomes more critical when the phase duration is short, as no node may be placed within the phase owing to the uniform time spacing of the constraints. Such dynamically infeasible motions can degrade the tracking performance of a controller~\cite{yoon2024spatio, zhao2024bi}.

In this work, to improve the consistency of translational dynamics feasibility, we propose a dynamically-consistent PBTO that maintains uniformly translational dynamics constraints throughout the entire trajectory, rather than only at points distributed evenly in time. 
Specifically, we transform the TO, for each contact point, into multiple-phase problems~\cite{rao2010algorithm}, which can be handled in a manner that allows adaptive adjustment of the constraint nodes based on the length of each phase, thereby ensuring consistent translational dynamic feasibility through the phase duration. To formulate the multiple-phase problem, we decouple the effects of forces generated at each contact point by leveraging the superposition property of translational dynamics.
Furthermore, we parameterize the separated forces and components of the robot's position as B{\'e}zier polynomials with time normalized by phase, and express their relationships through the analytical integration of the B{\'e}zier polynomials. By utilizing the two aforementioned methods, we ensure translational dynamics are satisfied across the trajectory.
Additionally, leveraging the fact that B{\'e}zier control points bound the curve, our method ensures the force profile always meets friction cone constraints.

Since angular dynamics are nonlinear, they cannot be handled via decomposition at individual contact points. Therefore, we enforce dynamics constraints at evenly spaced nodes. In spite of the constraints at evenly spaced nodes, the feasibility of angular dynamics is improved by considering the angular dynamics on \text{SO(3)} and reparameterize the robot's orientation as an orientation error in its tangent space during optimization \cite{hong2020real}. We believe this enables a more structured handling of angular dynamics while preserving the optimization framework.

The key contributions of this study are as follows:
\begin{itemize}

\item We proposed a method designed to be applicable within PBTO, ensuring the consistent satisfaction of translational dynamics. By utilizing contact point decomposition and the differentiation matrices of B{\'e}zier polynomials, we derived an analytical relationship between position and force.

\item In PBTO, we incorporate a method to ensure that the force profile always satisfies the friction cone constraints. This is achieved by leveraging the property that the B{\'e}zier polynomial remains within the convex hull formed by its control points.

\item We evaluated the trajectories generated by the proposed methodology in depth by introducing various scenarios and dynamic feasibility metrics. Compared to the baseline \cite{winkler2018gait}, the proposed method generates trajectories with improved dynamic feasibility across various environments.

\end{itemize}

\begin{figure}[t!]
      \centering
      \includegraphics[width=0.47\textwidth]{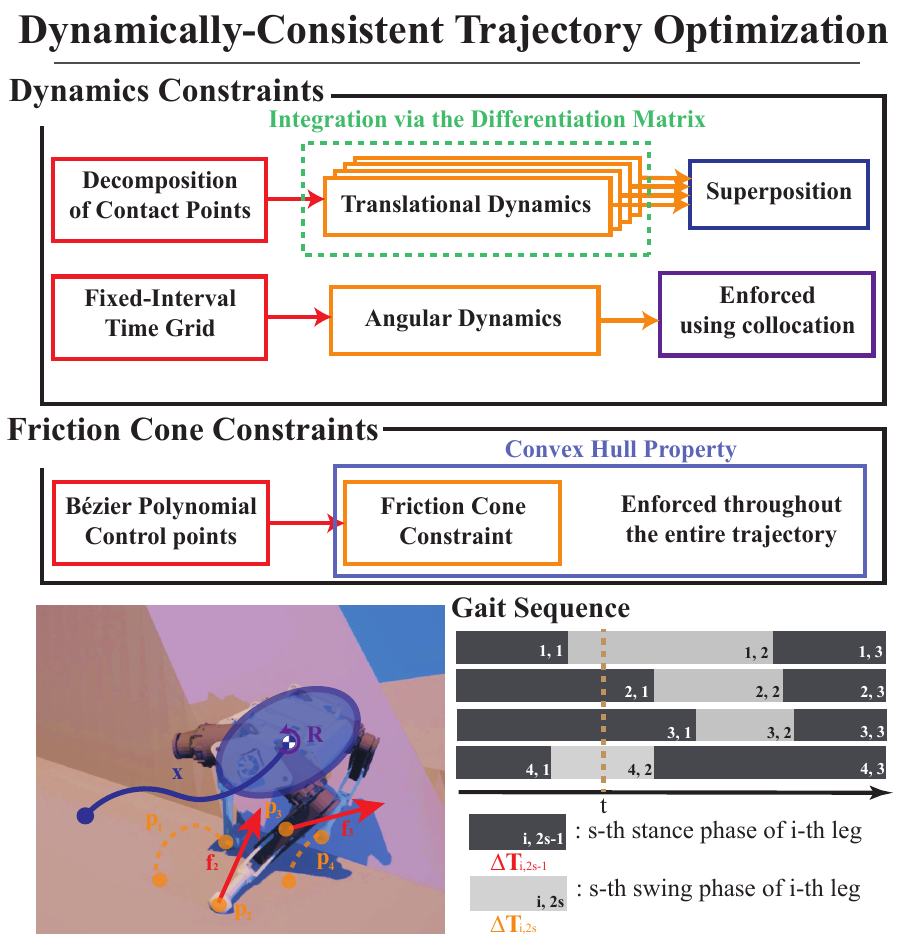}
      \caption{Overview of proposed method.}
      \label{fig:method_overview}
\end{figure}

\section{Preliminaries} \label{sec:preliminaries}

In this section, we provide background on our method. First, we introduce PBTO from \cite{winkler2018gait}. Then, we describe the system dynamics used in our approach. Finally, we summarize the definition and key properties of B{\'e}zier polynomials.

\subsection{Phase-Based Trajectory Optimization} \label{subsec:phase-based Method}

Given the initial state \( (\mathbf{X}_{\text{init}}, \dot{\mathbf{X}}_{\text{init}}) \), desired final state \( (\mathbf{X}_{\text{final}}, \dot{\mathbf{X}}_{\text{final}}) \), and total duration \( T_{\text{total}} \), PBTO generates body and foot trajectories, along with contact phase durations, for a legged robot. Since the contact sequence of a single leg consists of a sequential repetition of standing and swinging, adjusting the phase durations at each contact point enables the generation of any contact state at arbitrary times. Thus, PBTO incorporates phase durations as optimization variables, allowing the generation of arbitrary contact sequences.

The PBTO explicitly treats phase durations, allowing contact conditions to be handled implicitly. Since the force in the swing phase and the foot position in the stance phase are inherently determined by the phase itself, the method naturally ensures zero force during swing and a fixed foot during stance without additional constraints \cite{melon2020reliable}.

\subsection{System Dynamics} \label{subsec:system_dynamics}

We represent the robot using a reduced-order model, a Single Rigid Body (SRB) with point feet, neglecting leg mass.
The robot's state at time \(t\)  is defined as \((\mathbf{X}(t) := \{\mathbf{x}(t), \mathbf{R}(t), \mathbf{p}_1(t), \cdots, \mathbf{p}_{n_{\text{legs}}}(t)\}, \dot{\mathbf{X}}(t) := \{\dot{\mathbf{x}}(t), \boldsymbol{\omega}(t), \dot{\mathbf{p}}_1(t), \cdots, \dot{\mathbf{p}}_{n_{\text{legs}}}(t)\})\), where \(\mathbf{x}(t) \in \mathbb{R}^3\) and \(\mathbf{R}(t) \in \text{SO}(3)\) are the position and orientation of the body frame \(B\). \(\mathbf{p}_i(t) \in \mathbb{R}^3\) is the \(i\)-th foot's position. \(n_{\text{legs}}\) is the total number of legs.
%, which corresponds to the number of contactable points. 
\(\boldsymbol{\omega}(t) \in\mathbb{R}^3\) is the angular velocity expressed in \(B\). The control inputs are represented by the Ground Reaction Forces (GRF) \(\mathbf{f}_i(t) \in\mathbb{R}^3\) acting on each \(\mathbf{p}_i(t)\). All positions, orientations and forces, except for \(\boldsymbol{\omega}(t)\), are expressed in the inertial frame \(I\).

The robot's dynamics in this problem are given as follows:
\begin{subequations} \label{eq:dynamics}
\begin{align}
\ddot{\mathbf{x}} &= \frac{1}{m} \sum_i \mathbf{f}_i + \mathbf{g}, \label{eq:trans_dy}  \\
\dot{\mathbf{R}} &= \mathbf{R} \hat{\boldsymbol{\omega}}, \label{eq:ang_dy1}  \\
\dot{\boldsymbol{\omega}} &= \mathbf{I}^{-1} \left( \mathbf{R}^{\top} \left( \sum_i (\mathbf{p}_i-\mathbf{x}) \times \mathbf{f}_i \right) - \boldsymbol{\omega} \times (\mathbf{I} \mathbf{\omega}) \right), \label{eq:ang_dy2} \\
\mathbf{f}_i &\in F_i, \quad \forall i = 1, \ldots, n_{\text{legs}} ,\label{eq:cone_const} 
\end{align} 
\end{subequations}
where \(m\) is the robot's mass, \(\mathbf{g}\) is the gravity acceleration, and \(\mathbf{I}\) is the inertia matrix expressed in \(B\). \(\hat{\boldsymbol{\omega}} \in \mathbb{R}^{3\times3}\) is the skew-symmetric matrix representing the cross product for \(\omega\). 

\(F_i\) is expressed as a friction cone as follows:
\begin{align}
F_i = \left\{ \mathbf{f} \mid \|(\mathbb{I}-\mathbf{n}_i\mathbf{n}_i^{\top})\mathbf{f}\|_2 \leq \mu \mathbf{n}_i^{\top} \mathbf{f}, \right. \notag & \\  
\left. 0 \leq \mathbf{n}_i^{\top} \mathbf{f} \leq f^n_{\max} \right\}, \label{eq:cone}
\end{align}
where \(\mathbf{n}_i\) is the terrain's normal vector at \(\mathbf{p}_i\), \(\mu\) is the friction coefficient, and \(f^n_{\max}\) is the maximum normal force.

\subsection{Properties of B{\'e}zier Polynomials} \label{subsec:Bezier}

B{\'e}zier polynomials, a class of polynomials, are constructed as linear combinations of Bernstein polynomials. A B{\'e}zier polynomial of degree \( n \) is expressed as \( \mathbf{B}(s) = \sum_{k=0}^{n} \mathbf{c}_k b_k^n(s) \), where \( b_k^n(s) \) denotes the Bernstein polynomial of degree \( n \), defined as \( b_k^n(s) = \binom{n}{k} s^k (1 - s)^{n - k} \) for \( k = 0, \ldots, n \). The coefficients \( \mathbf{c}_k \) of the B{\'e}zier polynomial are called B{\'e}zier points or control points. 

According to the properties of the binomial function, \(\sum_{k=0}^{n} b_k^n(s) = 1\), and for \(0 \leq s \leq 1\), \(b_k^n(s) \geq 0\). Therefore, when \(a \leq t \leq b\), \(\mathbf{B}(\frac{t-a}{b-a})\) is a convex combination of \(\mathbf{C} := [\mathbf{c}_0 \ldots \mathbf{c}_n]^{\top}\), and if \(\mathbf{c}_k \in C\) for all \(k\), then \(\mathbf{B}(\frac{t-a}{b-a}) \in C\) \cite{prautzsch2002bezier}. Here, \(C\) is an arbitrary convex set.

Another useful property of B{\'e}zier polynomials is that taking the derivative of a B{\'e}zier polynomial results in another B{\'e}zier polynomial of lower degree. Moreover, the control points of the original polynomial and its derivative are related through a matrix transformation \cite{prautzsch2002bezier, li2020centroidal}. For example, the control points of \( \ddot{\mathbf{B}}\left(\frac{t-a}{b-a}\right) \), which is the second derivative of \( \mathbf{B}\left(\frac{t-a}{b-a}\right) \) with respect to time, are denoted as \(
\ddot{\mathbf{C}} := \begin{bmatrix} \ddot{\mathbf{c}}_0 & \ldots & \ddot{\mathbf{c}}_{n-2} \end{bmatrix}^{\top}
\). Using the second-order differentiation matrix \( \mathbf{K}^{n}_{b-a} \), \( \mathbf{C} \) can be expressed as follows:
\begin{align}
&\mathbf{K}^{n}_{b-a}\mathbf{C} = \begin{bmatrix}
    \ddot{\mathbf{C}} \\
    (\dot{\mathbf{B}}_{a})^{\top} \\
    (\mathbf{B}_{a})^{\top}
\end{bmatrix}, \notag \\
\mathbf{K}^{n}_{b-a} &= \begin{bmatrix}
    & \frac{n(n-1)}{(b-a)^2}\mathbf{\Phi}_{n} & \\
    -\frac{n}{b-a} & \frac{n}{b-a} & \textbf{0}_{n-1}^{\top} \\
    1 & 0 & \textbf{0}_{n-1}^{\top}
\end{bmatrix} \in \mathbb{R}^{n+1 \times n+1}, \label{eq:diff_matrix}
\end{align}
where \( \mathbf{\Phi}_{n} \) is an \( (n-1) \times (n+1) \) matrix with \(-2\) on the main diagonal, \(1\) on the adjacent diagonals, and zeros elsewhere. \( \mathbf{B}_{a} \) and \( \dot{\mathbf{B}}_{a} \) denote the initial values at \( t = a \).

\section{System Formulation}  \label{sec:Approach}

In this section, we explain a method to enforce translational dynamics and friction cone constraints by parameterizing it and reformulating the dynamics model.
 
\subsection{Trajectory Parameterization}  
\label{subsec:trajectory_parameterization}

We use the direct method to solve the TO problem by parameterizing \( \mathbf{p}_i \) and \( \mathbf{f}_i \) using B{\'e}zier polynomial bases of degree \( N \) and \( M \), respectively. For \( d \in \mathbb{N} \) and \( 2d-1 \leq n_{\text{phase}} \), where \( n_{\text{phase}} \) represents the number of phases for each leg, \( \mathbf{p}_i(t) \) and \( \mathbf{f}_i(t) \) are expressed using the control point sets  
\( \boldsymbol{\gamma}^d_i = [\boldsymbol{\gamma}^d_{(i,0)} \quad \boldsymbol{\gamma}^d_{(i,1)} \quad \cdots \quad \boldsymbol{\gamma}^d_{(i,N)}]^{\top} \)
and  
\( \boldsymbol{\alpha}^d_i = [\boldsymbol{\alpha}^d_{(i,0)} \quad \boldsymbol{\alpha}^d_{(i,1)} \quad \cdots \quad \boldsymbol{\alpha}^d_{(i,M)}]^{\top} \),  
respectively, as follows:
\begin{align}
\mathbf{p}_i(t) =
\begin{cases} 
\boldsymbol{\gamma}_{(i,0)}^d \, (= \mathbf{p}_i(\sigma_i^{2d-1})), & \sigma_i^{2d-2} \leq t < \sigma_i^{2d-1} \\
\sum_{k=0}^{N} \boldsymbol{\gamma}_{(i,k)}^d b_k^N(s^{\text{sw}}_{(i,d)}), & \sigma_i^{2d-1} \leq t < \sigma_i^{2d}
\end{cases}, \label{eq:p_bz}
\end{align}

\begin{align}
\mathbf{f}_i (t) =
\begin{cases} 
\sum_{k=0}^{M} \boldsymbol{\alpha}_{(i,k)}^d b_k^M(s^{\text{st}}_{(i,d)}), & \sigma_i^{2d-2} \leq t < \sigma_i^{2d-1} \\
\textbf{0}, & \sigma_i^{2d-1} \leq t < \sigma_i^{2d} 
\end{cases}, \label{eq:f_bz}
\end{align}
where the normalized time parameters are given by \( s^{\text{sw}}_{(i,d)} = \frac{t - \sigma_i^{2d-1}}{\Delta T_{i,2d}} \) and \( s^{\text{st}}_{(i,d)} = \frac{t - \sigma_i^{2d-2}}{\Delta T_{i,2d-1}} \). Here, \(\sigma_i^{l}\) is defined as:
\[
\sigma_i^{l} = \sum_{q=1}^{l} \Delta T_{i,q}. \label{eq:T_sum}
\]
%, \quad l \in \mathbb{Z}_{\geq 0}
The set of times when the \(i\)-th leg is in its \(d\)-th stance phase is denoted as \(C_{(i,d)} = \{t \mid 0 \leq t - \sigma_i^{2d-2} < \Delta T_{i,2d-1}\}\). In this article, we assume that all legs begin and end their sequences with a stance phase. Additionally, the total set of times when the \(i\)-th leg is in stance is denoted as \(C_i = \bigcup_d C_{(i,d)}\).

Unlike \(\mathbf{f}_i\) and \(\mathbf{p}_i\), \(\mathbf{R}\) and \(\boldsymbol{\omega}\) are discretized based on discrete-time \(t_k\) as \(\mathbf{R}_k\) and \(\boldsymbol{\omega}_k\) at constant intervals of \(\Delta t\). \(\mathbf{x}\) will be expressed in terms of the previously parameterized variables using the analytical translational dynamics relationship, as detailed in the next section.

\subsection{Restructuring of System Dynamics}  \label{subsec:reformulation_of_system_dynamics}

We reformulate dynamics \eqref{eq:dynamics} in terms of the parameterized variables defined in Section \ref{subsec:trajectory_parameterization}. In particular, for equations \eqref{eq:trans_dy} and \eqref{eq:cone_const}, we leverage the properties of B{\'e}zier polynomials discussed in Section \ref{subsec:Bezier} to ensure that they hold throughout the entire trajectory.

1) Translational dynamics: Translational dynamics \eqref{eq:trans_dy} is a linear second-order differential equation. Therefore, the robot's position \(\mathbf{x}\) can be divided into the position term \(\mathbf{x}^g\), determined by the initial conditions \((\mathbf{x}_{\text{init}}, \dot{\mathbf{x}}_{\text{init}})\) and \(\mathbf{g}\), and the position term \(\mathbf{x}^p_i\), which is caused solely by the \(i\)-th leg. 

The control input \(\mathbf{f}_i\) and the state \(\mathbf{x}^p_i\), divided by each contact point, share the same phase sequence, allowing the dynamics to be applied at consistent points for each phase rather than at fixed time intervals across the entire duration.
Building on this, we leverage the differential properties of B{\'e}zier polynomials discussed in Section \ref{subsec:Bezier} to analytically define the relationship between \(\mathbf{f}_i\) and \(\mathbf{x}^p_i\), providing a more precise characterization of their interaction.

When the \(i\)-th leg is in \(d\)-th stance phase, the B{\'e}zier polynomial \( \mathbf{x}^p_i \) is given in terms of \( \boldsymbol{\alpha}^d_i \) using \eqref{eq:diff_matrix}, where the control point set  
\( \boldsymbol{\beta}^d_i = [\boldsymbol{\beta}^d_{(i,0)} \quad \boldsymbol{\beta}^d_{(i,1)} \quad \cdots \quad \boldsymbol{\beta}^d_{(i,M+2)}]^{\top} \)  
is expressed as follows:
\begin{align}
\mathbf{x}^p_i(t) &=  \sum_{k=0}^{M+2} \boldsymbol{\beta}^d_{(i,k)} b_k^{M+2}(s^{\text{st}}_{(i,d)}), \label{eq:trans_int} \\
\boldsymbol{\beta}^d_i &=
(m\mathbf{K}^{M+2}_{\Delta T_{i,2d-1}})^{-1}\begin{bmatrix}
    \boldsymbol{\alpha}^d_i \\
    (\dot{\boldsymbol{\chi}}^{2d-1}_{i})^{\top} \\
    (\boldsymbol{\chi}^{2d-1}_{i})^{\top}
\end{bmatrix}, \label{eq:trans_int_matrix}
\end{align}
where \((\boldsymbol{\chi}^{2d-1}_i\), \(\dot{\boldsymbol{\chi}}^{2d-1}_i)\) are the initial state of the \((2d-1)\)-th segment of \(\mathbf{x}^p_i(t)\).
% while the $i$-th leg is in the $d$-th stance phase. $\alpha^d_{(i,k)}$ and $\beta^d_{(i,k)}$ are the $k$-th elements of $\alpha^d_i$ and $\beta^d_i$. 

On the other hand, when the \(j\)-th leg is in \(d'\)-th swing phase, $\mathbf{f}_j = \mathbf{0}$, so it can be simply represented as follows:
\begin{align}
\mathbf{x}^p_j(t) = \boldsymbol{\chi}^{2d'}_j + \dot{\boldsymbol{\chi}}^{2d'}_{j}(t-\sigma_i^{2d'-1}). \label{eq:swing_int}
\end{align}

Lastly, the complementary solution is a second-order polynomial due to \(\mathbf{g}\), \(\mathbf{x}_{\text{init}}\), \(\dot{\mathbf{x}}_{\text{init}}\), and is given by:
\begin{align}
\mathbf{x}^g(t) = \mathbf{x}_{\text{init}} + \dot{\mathbf{x}}_{\text{init}}t + \frac{1}{2}\mathbf{g}t^2. \label{eq:gravity_int}
\end{align}

In conclusion, \(\mathbf{x}(t)\) is given by the sum of the complementary solution \(\mathbf{x}^g\) and the particular solutions \(\mathbf{x}^p_i\) as follows:
\begin{align}
\mathbf{x}(t) = \mathbf{x}^g(t) + \sum_{i \in \{l \mid t \in C_l\}} \mathbf{x}^p_i(t) + \sum_{j \in \{l' \mid t \notin C_{l'}\}} \mathbf{x}^p_j(t). \label{eq:trans_int_sum} 
\end{align}

Using \eqref{eq:trans_int_sum}, we implicitly formulate the translational dynamics \eqref{eq:trans_dy} via direct shooting with analytical integration.

\(\mathbf{x}(t)\) is expressed solely in terms of the variable set \(\{\mathcal{A}, \Delta \mathcal{T}, \mathcal{X}, \mathcal{\dot{X}}\}\). Here, \(\mathcal{A}, \Delta \mathcal{T}, \mathcal{X}, \mathcal{\dot{X}}\) represent the sets of \(\boldsymbol{\alpha}^d_{(i,k)}, \Delta T_{i,q}, \boldsymbol{\chi}^{q}_i,\) and \(\dot{\boldsymbol{\chi}}^{q}_i\), respectively, which are related to \(\mathbf{x}(t)\). By using the fact that \(\boldsymbol{\chi}^{q}_i = \mathbf{x}^p_i(\sigma_i^{q-1})\) and \(\dot{\boldsymbol{\chi}}^{q}_i = \dot{\mathbf{x}}^p_i(\sigma_i^{q-1})\), \((\boldsymbol{\chi}^{q}_i, \dot{\boldsymbol{\chi}}^{q}_i)\) can be expressed as functions of \(\mathcal{A}\) and \(\Delta \mathcal{T}\), specifically as \(\boldsymbol{\chi}^{q}_i(\mathcal{A}, \Delta \mathcal{T})\) and \(\dot{\boldsymbol{\chi}}^{q}_i(\mathcal{A}, \Delta \mathcal{T})\). However, they are kept as variables to prevent the accumulation of nonlinearity, which is a common issue in the single shooting method~\cite{diehl2006fast}.
Additionally, since \(\mathbf{x}(t)\) is obtained analytically with respect to the variables, the required derivatives for optimization can also be computed analytically.

2) Angular dynamics: The dynamics \eqref{eq:ang_dy1} and \eqref{eq:ang_dy2} related to angular components are discretized at intervals of \(\Delta t\) as follows for \(1 \leq k \leq n_{\text{nodes}} - 1\):
\begin{align}
\mathbf{R}_{k+1} &= \mathbf{R}_k \text{Exp}(\boldsymbol{\omega}_k\Delta t), \label{eq:ang_int_1} \\
\boldsymbol{\omega}_{k+1} &= \boldsymbol{\omega}_{k} + \mathbf{I}^{-1} \Bigl( \mathbf{R}_k^{\top} (\sum_i (\mathbf{p}_i(t_k) - \mathbf{x}(t_k)) \times \mathbf{f}_i(t_k))  \notag \\
& \quad - \hat{\boldsymbol{\omega}}_k \mathbf{I} \boldsymbol{\omega}_k \Bigr) \Delta t, \label{eq:ang_int_2} 
\end{align}
where \(n_{\text{nodes}}\) is the number of discretized points. Exponential map Exp: \(\mathbb{R}^3 \to \text{SO}(3)\) maps the tangent space to the manifold. Logarithmic mapping Log: \(\text{SO}(3) \to \mathbb{R}^3\) is the inverse of the exponential map.

Since angular dynamics are nonlinear and each forcing function is not composed of a single phase sequence, it becomes challenging to decouple the dynamics for each contact. Therefore, we approached the problem in a collocation manner, as in the existing method \cite{winkler2018gait}, ensuring that the trajectory satisfies the dynamics at specific nodes spaced at fixed intervals throughout the entire time horizon. However, unlike \cite{winkler2018gait}, which represents rotations using a fixed local chart parameterized by Euler angles, we construct a local chart based on the tangent space at the current point on \(\text{SO}(3)\) during the optimization process. This approach allows for more robust optimization by addressing singularity issues and ensuring consistent representation \cite{hong2020real}.

%\subsubsubsection{Friction Cone}
3) Friction cone: When \(t \in C(i,d)\), the friction cone constraints \eqref{eq:cone_const} for the \(i\)-th leg are reformulated as follows:
\begin{align}
&\boldsymbol{\alpha}_{(i,k)}^d \in \Tilde{F}_i, \quad \forall k = 0, \ldots, M, \label{eq:reform_cone}\\
&\Tilde{F}_i = \left\{ \mathbf{f} \mid |\langle \mathbf{f}, \mathbf{t}^1_i \rangle| \leq \Tilde{\mu} \langle \mathbf{f}, \mathbf{n}_i \rangle, \right. \notag\\
&\qquad \qquad |\langle \mathbf{f}, \mathbf{t}^2_i \rangle| \leq \Tilde{\mu} \langle \mathbf{f}, \mathbf{n}_i \rangle, \notag\\
&\qquad \qquad \left. 0 \leq \langle \mathbf{f}, \mathbf{n}_i \rangle \leq \mathbf{f}^n_{\max} \right\} \notag, 
\end{align}
where \( \mathbf{t}^1_i \) and \( \mathbf{t}^2_i \) are the terrain's tangential vectors at \( \mathbf{p}_i \). \( \Tilde{F}_i \) is a friction pyramid, a linearized approximation of the friction cone with the effective friction coefficient \( \Tilde{\mu} \) (i.e., \( \Tilde{\mu} = \mu / \sqrt{2} \)).
Since \(\Tilde{F}_i\) consists of linear constraints, it helps improve numerical stability.

If \eqref{eq:reform_cone} is satisfied, then \( \mathbf{f}_i(t) \in \Tilde{F}_i \) for all \( t \), due to the relationship between the control points and the B{\'e}zier polynomial discussed in Section \ref{subsec:Bezier}.
Furthermore, since \(\Tilde{F}_i \subseteq F_i\), \eqref{eq:cone_const} is always satisfied. Although this is a conservative condition, increasing the number of control points, i.e., \(M\), can mitigate its limitations \cite{marcucci2023motion}.

\section{Trajectory Optimization}

Using the reformulated dynamics in Section \ref{subsec:reformulation_of_system_dynamics}, we express the optimization problem as follows:
\begin{align}
\begin{array}{ll}
\underset{\mathbf{\xi}}{\text{minimize}} & h(\mathbf{\xi}) = \sum_e w_e h_e(\mathbf{\xi}) \notag \\
\text{subject to} & \mathbf{c}_{\text{eq}}(\mathbf{\xi}) = \mathbf{0}, \notag \\
                  & \mathbf{c}_{\text{ineq}}(\mathbf{\xi}) \leq \mathbf{0} \notag,
\end{array}
\end{align}
where \( \mathbf{\xi} \) includes all B{\'e}zier control points for forces and foot positions, phase durations, discretized orientations, angular velocities, and initial states of the position segments. The objective function \( h(\mathbf{\xi}) \) is the weighted sum of costs \( h_e(\mathbf{\xi}) \) with weights \( w_e \). Equality and inequality constraints are denoted as \( \mathbf{c}_{\text{eq}}(\mathbf{\xi}) \) and \( \mathbf{c}_{\text{ineq}}(\mathbf{\xi}) \), respectively. These terms are defined in the remainder of this section.

\subsection{Objectives}
\hspace{0.5em}1) Nominal body trajectory:
Our method aims to ensure that the body path stably reaches the desired final state \((\mathbf{X}_{\text{final}}, \dot{\mathbf{X}}_{\text{final}})\) from the given initial state \((\mathbf{X}_{\text{init}}, \dot{\mathbf{X}}_{\text{init}})\). Therefore, we aim to guide the robot to follow the reference body trajectory as closely as possible while applying angular velocity regularization.
\begin{align}
h_1(\mathbf{\xi})&=\|\mathbf{x}^z(t_k)-z_{\text{ref}}(t_k)\|_2^2, \label{eq:z_obj} \\
h_2(\mathbf{\xi})&=\|\text{Log}(\mathbf{R}_{\text{ref}}(t_k)^{\top}\mathbf{R}_k)\|_2^2, \label{eq:R_obj} \\
h_3(\mathbf{\xi})&=\|\boldsymbol{\omega}_k\|_2^2 ,\label{eq:w_obj}\end{align}
where \((\cdot)^z\) denotes the z-component of the respective variable. The nominal body height \(z_{\text{ref}}(t_k)\) and rotation \(\mathbf{R}_{\text{ref}}\) are obtained via linear interpolation between the initial and final states over time in \(\mathbb{R}\) and \(\text{SO}(3)\), respectively.

2) Foot trajectory restrictions:
In the model, we consider massless point feet. Therefore, without appropriate restrictions, the optimized foot trajectory could become unrealistic. To address this, we ensure that the foot position remains close to a nominal trajectory, and that \(\dot{\mathbf{p}}_i\) and \(\ddot{\mathbf{p}}_i\) do not become excessively large.

\begin{align}
h_4(\mathbf{\xi})&=\|\mathbf{R}_k^{\top}(\mathbf{p}_i(t_{k})-\mathbf{x}(t_k))-\mathbf{p}^b_{\text{ref}, i}\|_2^2, \label{eq:p_obj} \\
h_5(\mathbf{\xi})&=\|\boldsymbol{\gamma}_{(i,k+2)}^d-2\boldsymbol{\gamma}_{(i,k+1)}^d+\boldsymbol{\gamma}_{(i,k)}^d\|_2^2, \label{eq:gamma_obj1} \\
h_6(\mathbf{\xi})&=\|\boldsymbol{\gamma}_{(i,k+1)}^d-\boldsymbol{\gamma}_{(i,k)}^d\|_2^2, \label{eq:gamma_obj2} 
\end{align}
where \( \mathbf{p}^b_{\text{ref}, i} \) is the nominal foot trajectory of the \(i\)-th leg relative to frame \(B\), which we simply treat as a fixed point.

Since $\mathbf{p}_i$ is parameterized by the control points $\boldsymbol{\gamma}_{(i,k)}^d$, as shown in \eqref{eq:p_bz}, the control points of $\ddot{\mathbf{p}}_i$ and $\dot{\mathbf{p}}_i$ are given by $\ddot{\boldsymbol{\gamma}}_{(i,k)}^d$ and $\dot{\boldsymbol{\gamma}}_{(i,k)}^d$, respectively, according to the differentiation properties of B{\'e}zier polynomials, as follows:
{\footnotesize
\begin{align*}
\ddot{\boldsymbol{\gamma}}_{(i,k)}^d &= \frac{N(N-1)\left(\boldsymbol{\gamma}_{(i,k+2)}^d - 2\boldsymbol{\gamma}_{(i,k+1)}^d + \boldsymbol{\gamma}_{(i,k)}^d\right)}{\left(\Delta T_{i,2d}\right)^2}, \\
\dot{\boldsymbol{\gamma}}_{(i,k)}^d &= \frac{N\left(\boldsymbol{\gamma}_{(i,k+1)}^d - \boldsymbol{\gamma}_{(i,k)}^d\right)}{\Delta T_{i,2d}}.
\end{align*}
}

Directly regularizing these expressions may lead to numerical instability as \(\Delta T_{i,2d} \to 0\). To avoid this, we regularize only the numerators, as shown in~\eqref{eq:gamma_obj1} and~\eqref{eq:gamma_obj2}.

\subsection{Constraints} \label {subsec:Constraints}
\hspace{0.5em}1) Final condition constraints:
The goal of this optimization problem is for the robot to reach \((\mathbf{X}_{\text{final}}, \dot{\mathbf{X}}_{\text{final}})\) at \(T_{\text{total}}\). Therefore, the trajectory must satisfy \(\mathbf{X}(T_{\text{total}}) = \mathbf{X}_{\text{final}}\) and \(\dot{\mathbf{X}}(T_{\text{total}}) = \dot{\mathbf{X}}_{\text{final}}\). Additionally, the trajectory must satisfy \(\sum_q \Delta T_{(i,q)} = T_{\text{total}}\).

2) Continuity constraints:
We introduced the initial state of the \(q\)-th segment of \(\mathbf{x}^p_i(t)\), \((\boldsymbol{\chi}^{q}_i, \dot{\boldsymbol{\chi}}^{q}_i)\), as optimization variables in the manner of multiple shooting.
To ensure the smoothness of \(\mathbf{x}^p_i (t)\), continuity constraints need to be added and are expressed as follows:
\begin{align}
& \mathbf{x}_i^p(\sigma^{q-1}_i) = \boldsymbol{\chi}^{q}_i, \,\dot{\mathbf{x}}_i^p(\sigma^{q-1}_i) = \dot{\boldsymbol{\chi}}^{q}_i. \label{eq:pos_eq} 
\end{align}

3) Foot motion constraints:
In this optimization formulation, we assume that there is no slip during foot contact. Therefore, when $t$ is within the stance phase, the foot must remain on the terrain and its velocity must be zero. 
\begin{align}
\boldsymbol{\gamma}_{(i,0)}^{(d,z)} &= h(\boldsymbol{\gamma}_{(i,0)}^{(d,x)},\boldsymbol{\gamma}_{(i,0)}^{(d,y)}), \label{eq:p_eq}\\
% \boldsymbol{\gamma}_{(i,1)}^{d} &= \boldsymbol{\gamma}_{(i,0)}^{d} = \boldsymbol{\gamma}_{(i,N)}^{d-1} = \boldsymbol{\gamma}_{(i,N-1)}^{d-1} \label{eq:gamma_eq}
\boldsymbol{\gamma}_{(i,0)}^{d} &= \boldsymbol{\gamma}_{(i,N)}^{d-1},
\end{align}
where \((\cdot)^x\), \((\cdot)^y\) denote the x-component and y-component of the respective variable, respectively. \( h((\cdot)^x, (\cdot)^y) \) represents the terrain height at \((\cdot)^x, (\cdot)^y\).
When $t_k$ is during the swing phase, the foot must not penetrate the terrain:
\begin{align}
\mathbf{p}^z_i(t_k) \geq h(\mathbf{p}_i^x(t_k), \mathbf{p}_i^y(t_k)), \quad \forall t_k \notin C_i, \label{eq:h_ineq}
\end{align}

Additionally, the distance between the $i$-th foot and the $i$-th hip must always be less than or equal to the maximum leg length $L_{\text{leg}}$ due to kinematic limitations:
\begin{align}
\|\mathbf{R}_k^{\top}(\mathbf{p}_i(t_k) - \mathbf{x}(t_k)) - \mathbf{x}^{b}_{\text{hip}, i}\|_2 \leq L_{\text{leg}},
\label{eq:p_ineq}
\end{align}
where \(\mathbf{x}^b_{\text{hip}, i}\) is a constant representing the hip position of the \(i\)-th leg relative to frame \(B\).

4) Friction cone constraints:
To ensure that the force profile satisfies Coulomb's law, we add \eqref{eq:reform_cone} to the constraints.

5) Angular dynamics constraints:
We add \eqref{eq:ang_int_1} and \eqref{eq:ang_int_2} to the constraints to ensure that the angular dynamics are guaranteed only at the discretized time instances.

\subsection{Solver}
We solve our TO problem using sequential quadratic programming (SQP), which approximates the original problem as a series of quadratic programs (QP). For the QP solver, we utilize qpSWIFT \cite{pandala2019qpswift}.
To prevent computational inefficiencies in the framework, analytical gradients are used for the QP approximation. 
The Hessian is replaced with a Gauss-Newton Hessian approximation.

The SQP process is terminated when the following conditions are satisfied, indicating convergence:

{\small
\begin{align}
&\frac{h(\mathbf{\xi}_{u-1}) - h(\mathbf{\xi}_{u})}{h(\mathbf{\xi}_{u})} < 0.1, \label{eq:obj_conv} \\
&\frac{\phi_{\text{eq}}(\mathbf{\xi}_{u-1}) - \phi_{\text{eq}}(\mathbf{\xi}_{u})}{\phi_{\text{eq}}(\mathbf{\xi}_{u})} < 0.1, \quad \phi_{\text{eq}}(\mathbf{\xi}_{u}) < 0.001 \label{eq:eq_conv}, \\
&\frac{\phi_{\text{ineq}}(\mathbf{\xi}_{u-1}) - \phi_{\text{ineq}}(\mathbf{\xi}_{u})}{\phi_{\text{ineq}}(\mathbf{\xi}_{u})} < 0.1, \quad \phi_{\text{ineq}}(\mathbf{\xi}_{u}) < 0.001, \label{eq:ineq_conv}
\end{align}
}where \(\mathbf{\xi}_u\) denotes the value of the optimization variable at the \(u\)-th SQP iteration. 
The constraint feasibilities, \(\phi_{\text{eq}}\) and \(\phi_{\text{ineq}}\), are defined as follows, 
with \(n_{\mathbf{c}_{\text{eq}}}\) and \(n_{\mathbf{c}_{\text{ineq}}}\) denoting the numbers of equality 
and inequality constraints, respectively:

\begin{align}
\phi_{\text{eq}}(\mathbf{\xi}_{u}) = \frac{\| \mathbf{c}_{\text{eq}}(\mathbf{\xi}_u)\|_1}{n_{\mathbf{c}_{\text{eq}}}}, \quad 
\phi_{\text{ineq}}(\mathbf{\xi}_{u}) = \frac{\| (\mathbf{c}_{\mathrm{ineq}}(\boldsymbol{\xi}_u))_{+} \|_1}{n_{\mathbf{c}_{\text{ineq}}}}, \label{eq:conv_con}
\end{align}
where $(\cdot)_+$ denotes the elementwise positive part.

\begin{figure}[t!]
      \centering
      \includegraphics[width=0.47\textwidth]{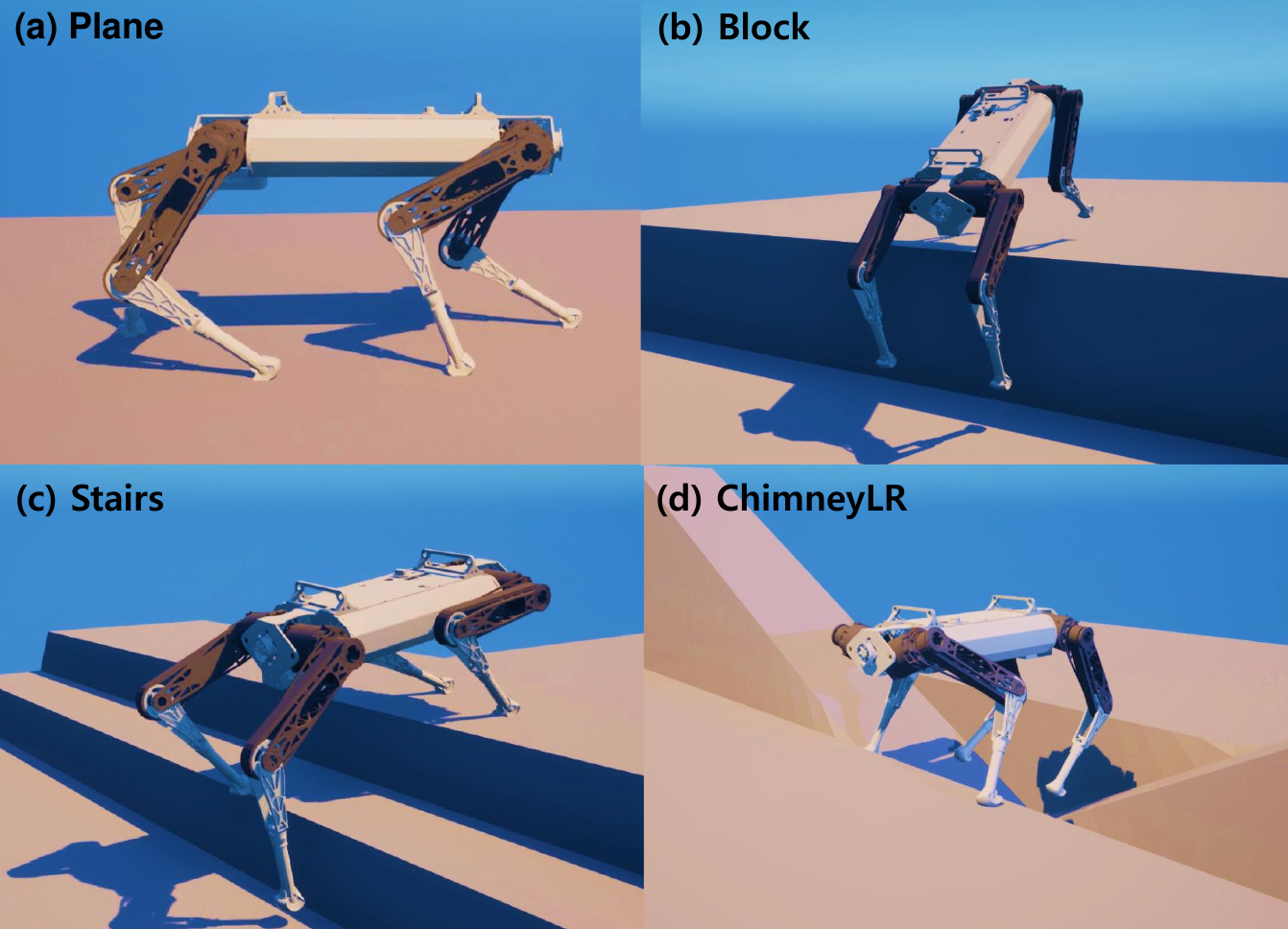}
      \caption{The proposed TO framework generated motion on various terrains. \textbf{(a) Plane} is a flat surface, \textbf{(b) Block} represents a 0.5m step, \textbf{(c) Stairs} is a 2-step terrain, and \textbf{(d) ChimneyLR} represents continuous vertical walls inclined at 45 degrees. All robot visualizations were generated using RaiSimUnreal \cite{hwangbo2018per}.}
      \label{fig:various_terrain}
\end{figure}

\begin{table*}[]
\centering
\captionsetup{justification=centering}  
\caption{Number of Optimization Failures and The Mean Values (And Standard Deviation) of Dynamics and Friction Cone Constraints Violation for our method, our method without the Convex Hull Property (CHP), and the baseline.}
\resizebox{1\textwidth}{!}{
\begin{tabular}{r|ccc|ccc}
\hline
\multicolumn{1}{c|}{\multirow{2}{*}{}}        & \multicolumn{3}{c|}{(a) Plane}                                 & \multicolumn{3}{c}{(b) Block}                                   \\ \cline{2-7} 
\multicolumn{1}{c|}{}                         & Ours                  & Ours (w/o CHP)       & Baseline        & Ours                   & Ours (w/o CHP)       & Baseline        \\ \hline
\multicolumn{1}{c|}{Failure Count}            & \textbf{0}            & \textbf{0}           & \textbf{0}      & 4                      & \textbf{3}           & \textbf{3}      \\ \hline
\multicolumn{1}{c|}{TD Violation x {[}N{]}}   & \textbf{0 (0)}        & \textbf{0 (0)}       & 62.34 (132.04)  & \textbf{0 (0)}         & \textbf{0 (0)}       & 75.28 (179.52)  \\
y                                             & \textbf{0 (0)}        & \textbf{0 (0)}       & 37.94 (87.42)   & \textbf{0 (0)}         & \textbf{0 (0)}       & 43.49 (110.60)  \\
z                                             & \textbf{0 (0)}        & \textbf{0 (0)}       & 163.97 (380.61) & \textbf{0 (0)}         & \textbf{0 (0)}       & 178.54 (447.32) \\ \hline
\multicolumn{1}{c|}{AD Violation x {[}N m{]}} & 5.77 (2.18)           & \textbf{5.76 (2.67)} & 24.25 (60.26)   & \textbf{6.66 (2.47)}   & 6.66 (3.15)          & 27.96 (92.77)   \\
y                                             & \textbf{11.11 (2.31)} & 11.56 (2.26)         & 37.98 (93.80)   & \textbf{13.35 (2.84)}  & 13.42 (3.00)         & 42.79 (125.12)  \\
z                                             & 2.70 (1.19)           & \textbf{2.50 (1.07)} & 11.01 (25.43)   & 3.28 (1.60)            & \textbf{3.05 (1.56)} & 14.04 (37.31)   \\ \hline
\multicolumn{1}{c|}{FC Violation FR {[}N{]}}  & \textbf{0 (0)}        & 0.07 (0.11)          & 42.21 (144.07)  & \textbf{0 (0)}         & 0.08 (0.12)          & 49.84 (195.14)  \\
FL                                            & \textbf{0 (0)}        & 0.07 (0.11)          & 41.78 (151.47)  & \textbf{2e-8 (3e-7)}   & 0.08 (0.12)          & 55.51 (198.83)  \\
RR                                            & \textbf{0 (0)}        & 0.07 (0.12)          & 45.07 (158.92)  & \textbf{2e-4 (1e-3)}   & 0.08 (0.12)          & 53.10 (186.16)  \\
RL                                            & \textbf{0 (0)}        & 0.07 (0.13)          & 47.01 (163.95)  & \textbf{3e-3 (5e-2)}   & 0.09 (0.12)          & 49.21 (195.28)  \\ \hline
\multicolumn{1}{c|}{\multirow{2}{*}{}}        & \multicolumn{3}{c|}{(c) Stairs}                                & \multicolumn{3}{c}{(d) ChimneyLR}                               \\ \cline{2-7} 
\multicolumn{1}{c|}{}                         & Ours                  & Ours (w/o CHP)       & Baseline        & Ours                   & Ours (w/o CHP)       & Baseline        \\ \hline
\multicolumn{1}{c|}{Failure Count}            & 8                     & \textbf{1}           & 17              & 4                      & \textbf{1}           & 0               \\ \hline
\multicolumn{1}{c|}{TD Violation x {[}N{]}}   & \textbf{0 (0)}        & \textbf{0 (0)}       & 82.77 (139.44)  & \textbf{0 (0)}         & \textbf{0 (0)}       & 38.15 (98.68)   \\
y                                             & \textbf{0 (0)}        & \textbf{0 (0)}       & 51.12 (92.69)   & \textbf{0 (0)}         & \textbf{0 (0)}       & 21.21 (58.95)   \\
z                                             & \textbf{0 (0)}        & \textbf{0 (0)}       & 252.11 (432.61) & \textbf{0 (0)}         & \textbf{0 (0)}       & 112.65 (337.52) \\ \hline
\multicolumn{1}{c|}{AD Violation x {[}N m{]}} & 7.40 (2.11)           & \textbf{7.29 (2.43)} & 32.88 (63.41)   & \textbf{6.03 (2.17)}   & 6.06 (2.38)          & 11.48 (34.93)   \\
y                                             & \textbf{12.33 (2.42)} & 12.87 (2.78)         & 55.44 (102.87)  & \textbf{12.12 (2.15)}  & 12.53 (2.06)         & 19.44 (54.84)   \\
z                                             & 3.96 (1.58)           & \textbf{3.62 (1.58)} & 14.93 (24.81)   & 2.98 (1.30)            & \textbf{2.79 (1.07)} & 7.77 (26.19)    \\ \hline
\multicolumn{1}{c|}{FC Violation FR {[}N{]}}  & \textbf{0 (0)}        & 0.12 (0.14)          & 67.80 (172.60)  & \textbf{5e-14 (8e-13)} & 0.07 (0.12)          & 21.81 (103.69)  \\
FL                                            & \textbf{0 (0)}        & 0.12 (0.14)          & 70.18 (180.96)  & \textbf{3e-6 (5e-5)}   & 0.08 (0.11)          & 20.55 (91.61)   \\
RR                                            & \textbf{0 (0)}        & 0.13 (0.14)          & 67.02 (173.45)  & \textbf{0 (0)}         & 0.07 (0.11)          & 21.63 (100.50)  \\
RL                                            & \textbf{0 (0)}        & 0.13 (0.15)          & 67.84 (180.02)  & \textbf{8e-8 (1e-6)}   & 0.09 (0.13)          & 19.43 (97.80)   \\ \hline
\end{tabular}
}
\label{tab:feasibility}
\end{table*}

\section{RESULTS}
In this section, we evaluate our TO framework by comparing the results obtained using the quadruped robot with the baseline. Here, the baseline employs the open-source TO library TOWR \cite{winkler2018towr}, which implements a TO framework based on the collocation method \cite{winkler2018gait}, where dynamics are constrained at evenly spaced time intervals.

\begin{figure}[t!]
    \centering
    \includegraphics[width=0.48\textwidth]{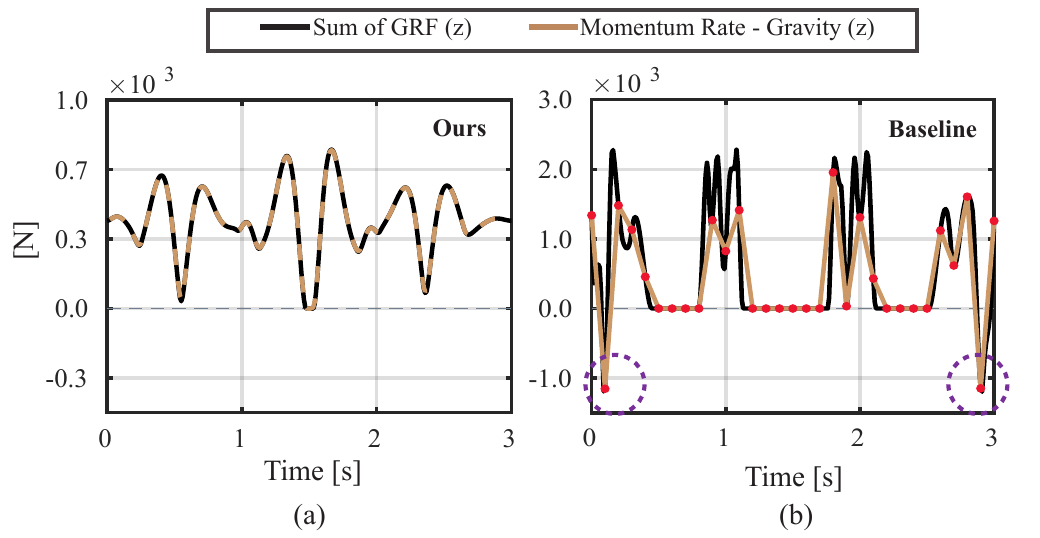}
    \caption{
    The gravity-compensated linear momentum derivative and total GRF along the z-axis for trajectories generated by the proposed method and the baseline when the robot bounded 3 meters on a plane with $\tilde{\mu} = 1$. Red dots in (b) indicate the nodes where the translational dynamics are explicitly satisfied. In (a), we numerically calculated the linear momentum derivative by taking the second derivative of \( m \mathbf{x}(t) \), demonstrating the feasibility of the continuity constraints.
    The initial and final base positions are $\mathbf{x}_{\text{init}} = [0 \text{m},\ 0 \text{m},\ 0.51 \text{m}]^{\top}$ and $\mathbf{x}_{\text{fin}} = [3 \text{m},\ 0 \text{m},\ 0.51 \text{m}]^{\top}$. The initial and final linear and angular base velocities are zero, and the base orientation is aligned with the world frame, i.e., the rotation matrix is identity at both ends.
    }
    \label{fig:linear_z_graph}
\end{figure}

\subsection{Trajectory Validation}

Our proposed TO framework is capable of operating under various conditions, and the motion generated on different terrains is shown in Fig.~\ref{fig:various_terrain}. 

To measure the computation duration and dynamics feasibility of the proposed TO, we generated 300 samples under different conditions for each terrain depicted in Fig.~\ref{fig:various_terrain}. For each terrain, the robot was required to traverse a distance of 3 meters in 3 seconds. The initial and final x-axis velocities were sampled uniformly from \([-0.5 \, \text{m/s}, 0.5 \, \text{m/s}]\), the initial and final yaw angles from \([-45^\circ, 45^\circ]\), and the effective friction coefficient from \([0.6, 1.0]\). Additionally, the initial values of the gait duration, which are optimization variables, were randomly sampled. The baseline samples were generated under the same conditions. The TO was performed on a desktop PC with an AMD Ryzen 7 5800X CPU. 

For our TO problems, \(n_{\text{nodes}}\) was set to 30, and \(n_{\text{phase}}\) was set to 7, resulting in a total of 1,216 optimization variables \(n_z\). The number of equality constraints \(n_{\mathbf{c}_{\text{eq}}}\) and inequality constraints \(n_{\mathbf{c}_{\text{ineq}}}\) were 836 and 976, respectively. In the baseline TO problem, \(n_{\text{phase}}\) also was set to 7, and \(n_z = 708\). \(n_{\mathbf{c}_{\text{eq}}}\) and \(n_{\mathbf{c}_{\text{ineq}}}\) were 442 and 1,076, respectively. We selected \(n_{\text{phase}}\) based on the total horizon \(T_{\text{total}}\) and the reasonable phase duration \(\Delta T = \, 0.5\,\text{s}\), with \(n_{\text{phase}} \approx \frac{T_{\text{total}}}{\Delta T}\).

Table~\ref{tab:feasibility} shows the mean and standard deviation of dynamics feasibility for each terrain. The mean and standard deviation were evaluated only for cases where the optimizer did not fail. The criteria for determining failure are as follows: in our case, failure is determined if the SQP iterations do not satisfy \eqref{eq:obj_conv}, \eqref{eq:eq_conv}, and \eqref{eq:ineq_conv} within 100 iterations. For the baseline, failure is defined as the case when IPOPT \cite{wachter2006implementation} returns a failure flag.

We define Translational Dynamics Violation (TD Violation), Angular Dynamics Violation (AD Violation), and Friction Cone Violation (FC Violation) as follows to analyze the dynamics satisfaction of the trajectory generated by the TO framework:
{
\begin{align}
&\text{TD Violation}=\frac{1}{T_{\text{total}}}\sum \frac{\text{TD}(\tau_{k+1}) + \text{TD}(\tau_k)}{2}\Delta \tau, \label{eq:TD_feas}\\
&\quad \textnormal{TD} = |m \ddot{\mathbf{x}} - m\mathbf{g} - \sum_i \mathbf{f}_i| \notag,
\end{align}
\begin{align}
&\text{AD Violation}=\frac{1}{T_{\text{total}}}\sum \frac{\text{AD}(\tau_{k+1}) + \text{AD}(\tau_k)}{2}\Delta \tau, \label{eq:AD_feas}\\
&\quad \text{AD} = 
|\mathbf{R}(\mathbf{I}\dot{\boldsymbol{\omega}} + \boldsymbol{\omega} \times \mathbf{I} \boldsymbol{\omega}  -  \sum_i (\mathbf{p}_i-\mathbf{x}) \times \mathbf{f}_i| \notag,
\end{align}
\begin{align}
\text{FC Violation}_i &= \frac{1}{\int_{\tau \in C_i} d\tau} \sum \frac{ \text{FC}_i(\tau_{k+1}) + \text{FC}_i(\tau_k)}{2} \Delta \tau, \label{eq:FC_feas}\\
&\text{FC}_i = \min_{\tilde{\mathbf{f}} \in \tilde{F}_i} \left\| \mathbf{f}_i - \tilde{\mathbf{f}} \right\|_2 \notag,
\end{align}
}where time is discretized into intervals of \(\Delta \tau\) and denoted by \(\tau_k\), with \(\Delta \tau\) set to \(0.01 \, \text{s}\). \(\tilde{F}_i\) represents the friction pyramid as defined in \eqref{eq:reform_cone}, and \(\text{FC}_i\) denotes the Euclidean distance from \(\mathbf{f}_i\) to the closest point on the friction pyramid. Violations represent the average dynamics violations over time, computed using Trapezoidal Approximations for numerical integration. Note that, due to the discretized angular dynamics used in the proposed framework, \(\dot{\boldsymbol{\omega}}(t)\) is calculated as the constant value \((\boldsymbol{\omega}_{k+1} - \boldsymbol{\omega}_k) / \Delta t\), where \(t_k \leq t < t_{k+1}\).

Since our method uses analytical integration of \(\mathbf{x}(t)\) with respect to \(\mathbf{f}_i(t)\), TD Violation is always \textbf{0}. Additionally, as the control points of the B{\'e}zier curve for \(\mathbf{f}_i\) are constrained within the friction cone, FC Violation \(\approx \textbf{0}\), except for numerical errors, as shown in Table\ \ref{tab:feasibility} due to the properties of the B{\'e}zier curve. Furthermore, AD Violation is also smaller than the baseline, which can be attributed to our consistent \(\text{SO}(3)\) representation.

The proposed method required a mean and standard deviation of 7.283~\si{\second} and 6.350~\si{\second} to solve, compared to 5.936~\si{\second} and 4.836~\si{\second} for the baseline. Optimization failures were also more frequent in scenarios (b) and (d). While our framework improves dynamics and friction cone feasibility, it incurs slightly higher computation time and, in some cases, more failures. This is because, unlike the baseline, which allows feasibility violations along the trajectory, our method must operate within a more constrained space, strictly enforcing translational dynamics and friction cone conditions throughout the entire trajectory.

As in the baseline, we applied the friction cone constraint only at specific nodes, while the translational dynamics were enforced over the entire trajectory and the orientation was represented using $\mathrm{SO}(3)$, as in our formulation. This reduced the mean and standard deviation of solving time to 5.30~\si{\second} and 3.80~\si{\second}, respectively, resulting in faster computation than the baseline. As shown in Table~\ref{tab:feasibility}, optimization failures also decreased across all terrains. Despite the relaxation, violations of dynamics and friction cone feasibility remained significantly lower than those of the baseline.

Fig.~\ref{fig:linear_z_graph} compares the total GRF and the gravity-compensated linear momentum derivative along the z-axis for trajectories generated by \textbf{(a)} Ours and \textbf{(b)} the Baseline. While \textbf{(a)} satisfies the dynamics throughout the entire trajectory, \textbf{(b)} enforces the translational dynamics only at discrete time nodes (red dots). As a result, in \textbf{(b)}, the momentum derivative fails to capture the local maxima of the total GRF.

\begin{figure}[t!]
      \centering
      \includegraphics[width=0.47\textwidth]{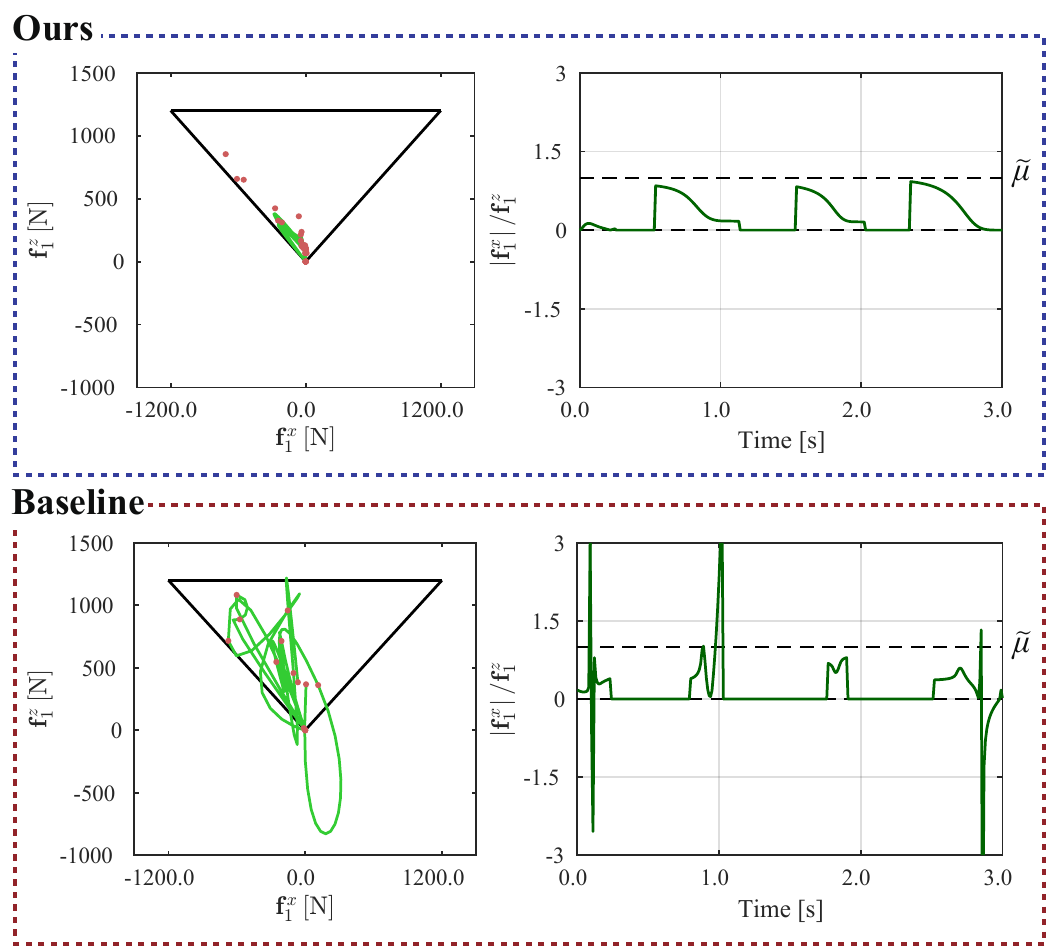}
      \caption{GRF profile projected onto the x-z plane (left) and x-directional force magnitude relative to the z-directional force over time (right) for the front right leg in the proposed method and the baseline. In the left graph, the black line represents the friction pyramid projected onto the x-z plane, while the red dots indicate the control points and the points where the friction cone constraints are applied in the proposed method and baseline.}
    \label{fig:friction_cone}
\end{figure}

In \textbf{(b)}, the purple dot circles indicate instances where the sum of the GRF becomes negative, caused by deviations of the GRF trajectory from the friction pyramid. This issue arises because, in the baseline, friction cone constraints are enforced only at fixed intervals in each phase, rather than continuously. Consequently, translational dynamics are sometimes applied to physically infeasible forces.

Fig.~\ref{fig:friction_cone} further illustrates the GRF profile projected onto the x-z plane of the front right leg, along with the x-directional force magnitude and the z-directional force ratio over time. Since our method enforces constraints on the control points of the B{\'e}zier polynomial, the trajectory remains within the friction pyramid throughout. In contrast, the baseline's interval-based enforcement leads to instances where the GRF significantly deviates from the friction pyramid.

\begin{table}[t]
\centering
\caption{Tracking Error with Respect to Reference Trajectories Generated by Our Method and the Baseline. Mean Values (and Standard Deviation).}
\resizebox{0.48\textwidth}{!}{
\begin{tabular}{l|cc|cc}
\hline
\multirow{2}{*}{} & \multicolumn{2}{c|}{Position Tracking Error [m]} & \multicolumn{2}{c}{Orientation Tracking Error [rad]} \\
\cline{2-5}
                  & Ours             & Baseline        & Ours               & Baseline        \\
\hline
X                 & \textbf{0.132 (0.036)} & 0.165 (0.052)   & \textbf{0.061} (0.041) & 0.104 (0.046) \\
Y                 & \textbf{0.048 (0.031)} & 0.069 (0.035)   & \textbf{0.028} (0.008) & 0.047 (0.009) \\
Z                 & \textbf{0.016 (0.002)} & 0.024 (0.002)   & \textbf{0.065} (0.024) & 0.111 (0.049) \\
\hline
\end{tabular}
}
\label{tab:tracking_error}
\end{table}

\subsection{Trajectory Tracking Validation}

Our trajectory optimization framework can be used as a reference trajectory generator for model-based controllers. To evaluate how well the controller can track the generated motions, we assessed the performance of both the proposed framework and the baseline using a model predictive controller (MPC)~\cite{di2018dynamic} on a quadruped robot. A total of 50 simulations were conducted, and the tracking errors in position and orientation were computed. The tracking error is defined as the time-averaged absolute difference between the reference trajectory and the simulated robot trajectory.

The evaluation was performed under randomized conditions, where the initial and final yaw angles were sampled from the range $[-15^\circ,\ 15^\circ]$, and the effective friction coefficient was sampled from $[0.5,\ 1.0]$. Each task required the robot to traverse a distance of 1.5 meters within 3 seconds.

As shown in Table~\ref{tab:tracking_error}, the trajectories generated by the proposed framework resulted in improved tracking performance. This suggests that the proposed framework produces trajectories that are more suitable for the controller and more consistent with both the system dynamics and the friction cone constraints.

\section{Conclusion}

In this paper, we presented a novel PBTO framework for legged robots, ensuring satisfaction of translational dynamics and friction cone constraints across the trajectory. Specifically, we utilized contact point decomposition and the differentiation matrices of B{\'e}zier polynomial to guarantee translational dynamics and leveraged the convex hull property of B{\'e}zier curves to enforce friction cone constraints. Moreover, we consider the angular dynamics on \text{SO(3)} and reparameterize the robot's orientation as an orientation error in the tangent space during optimization, leading to the reduction in AD violation.

The proposed TO framework demonstrated the capability to generate diverse dynamics motions and showed improved dynamic consistency over previous methods that rely solely on the collocation approach \cite{winkler2018gait}, where dynamics are enforced at fixed time steps.
 
In future work, we plan to utilize our dynamically consistent TO framework as a reference motion generator for model-based controllers \cite{ahn2021versatile} and as a target motion generator for learning-based controllers \cite{brakel2022learning, wu2023learning}, in scenarios where dynamics and friction cone feasibility are critical. Our TO framework is capable of generating feasible solutions over a variety of terrain conditions.

\bibliographystyle{IEEEtran}
\bibliography{ref}   % <- ref.bib
% \renewcommand{\bibfont}{\footnotesize}
% \printbibliography


%\vfill


\end{document}